\documentclass[conference]{IEEEtran}
\IEEEoverridecommandlockouts
\usepackage{cite}
\usepackage{multirow,subcaption}
\usepackage{amsmath,amssymb,amsfonts}
\usepackage{algorithm,algorithmic}
\usepackage{graphicx}
\usepackage{textcomp,balance}
\usepackage{xcolor,todonotes,booktabs,url}
\def\BibTeX{{\rm B\kern-.05em{\sc i\kern-.025em b}\kern-.08em
    T\kern-.1667em\lower.7ex\hbox{E}\kern-.125emX}}

\begin{document}

\title{Experience-Driven PCG via Reinforcement Learning: A Super Mario Bros Study
\thanks{This paper is accepted by the 2021 IEEE Conference on Games and will be included in the proceedings.}
}

\author{\IEEEauthorblockN{Tianye Shu\IEEEauthorrefmark{1}\IEEEauthorrefmark{2}, Jialin Liu\IEEEauthorrefmark{1}\IEEEauthorrefmark{2}}
\IEEEauthorblockA{\IEEEauthorrefmark{1} \textit{Research Institute of Trustworthy Autonomous System}\\
\textit{Southern University of Science and Technology}}
\IEEEauthorblockA{\IEEEauthorrefmark{2}\textit{Guangdong Provincial Key Laboratory of Brain-inspired Intelligent Computation}\\
\textit{Department of Computer Science and Engineering} \\
\textit{Southern University of Science and Technology}\\
Shenzhen, China\\
11710101@mail.sustech.edu.cn, liujl@sustech.edu.cn}
\and
\IEEEauthorblockN{Georgios N. Yannakakis}
\IEEEauthorblockA{\textit{Institute of Digital Games} \\
\textit{University of Malta}\\
Msida, Malta}
georgios.yannakakis@um.edu.mt
}

\IEEEoverridecommandlockouts
\IEEEpubid{\makebox[\columnwidth]{978-1-6654-3886-5/21/\$31.00~\copyright2021 IEEE\hfill} \hspace{\columnsep}\makebox[\columnwidth]{ }}
\maketitle
\IEEEpubidadjcol

\begin{abstract}
We introduce a procedural content generation (PCG) framework at the intersections of experience-driven PCG and PCG via reinforcement learning, named ED(PCG)RL, EDRL in short. EDRL is able to teach RL designers to generate endless playable levels in an online manner while respecting particular experiences for the player as designed in the form of reward functions. The framework is tested initially in the Super Mario Bros game. In particular, the RL designers of Super Mario Bros generate and concatenate level segments while considering the diversity among the segments. The correctness of the generation is ensured by a neural net-assisted evolutionary level repairer and the playability of the whole level is determined through AI-based testing. Our agents in this EDRL implementation learn to maximise a quantification of Koster's principle of \emph{fun} by moderating the degree of diversity across level segments. Moreover, we test their ability to design fun levels that are diverse over time and playable. Our proposed framework is capable of generating endless, playable Super Mario Bros levels with varying degrees of fun, deviation from earlier segments, and playability. EDRL can be generalised to any game that is built as a segment-based sequential process and features a built-in compressed representation of its game content. 
\end{abstract}
\begin{IEEEkeywords}
PCGRL, EDPCG, online level generation, procedural content generation, Super Mario Bros
\end{IEEEkeywords}

\section{Introduction}\label{sec:intro}

Procedural content generation (PCG)~\cite{shaker2016procedural,yannakakis2018artificial} is the algorithmic process that enables the (semi-)autonomous design of games to satisfy the needs of designers or players. As games become more complex and less linear, and uses of PCG tools become more diverse, the need for generators that are reliable, expressive, and trustworthy is increasing. Largely speaking, game content generators can produce outcomes either in an offline or in an online manner \cite{yannakakis2018artificial}. Compared to offline PCG, online PCG is flexible, dynamic and interactive but it comes with several drawbacks: it needs to be able to generate meaningful content rapidly without causing any catastrophic failure to the existing game content. Because of the many challenges that arise when PCG systems operate online (i.e. during play), only limited studies have focused on that mode of generation~\cite{shaker2010towards,jennings2010polymorph,stammer2015player,shi2017learning}. 

In this paper we introduce a framework for online PCG at the intersections of the experience-driven PCG (EDPCG) \cite{yannakakis2011experience} and the PCG via reinforcement learning (PCGRL)~\cite{khalifa2020pcgrl} frameworks. The ED(PCG)RL framework, EDRL for short, enables the generation of personalised content via the RL paradigm. EDRL builds upon and extends PCGRL as it makes it possible to generate endless levels of arcade-like games beyond the General Video Game AI (GVGAI) framework \cite{perez2019general,gvgaibook2019} in an online fashion. It also extends the EDPCG framework as it enables RL agents to create personalised content that is driven by experience-based reward functions. EDRL benefits from experience-driven reward functions and RL agents that are able to design in real-time based on these functions. The result is a versatile generator that can yield game content in an online fashion respecting certain functional and aesthetic properties as determined by the selected reward functions.

We test EDRL initially in \emph{Super Mario Bros} (SMB) (Nintendo, 1985)\cite{nintendo1985super} through the generative designs of RL agents that learn to optimise certain reward functions relevant to level design. In particular, we are inspired by Koster's \emph{theory of fun} \cite{koster2013theory} and train our RL agents to moderate the level of \emph{diversity} across SMB level segments. Moreover, we test the notion of \emph{historical deviation} by considering earlier segment creations when diversifying the current segment. Finally, we repair defects in levels (e.g., broken pipes) via a neural net-assisted evolutionary repairer~\cite{shu2020cnet}, and then check the playability of levels through agent-based testing. Importantly, EDRL is able to operate online in SMB as it represents the state and action via a latent vector. The key findings of the paper suggest that EDRL is possible in games like SMB; the RL agents are able to online generate playable levels of varying degrees of \emph{fun} that deviate over time.

Beyond introducing the EDRL framework, we highlight a number of ways this paper contributes to the current state of the art.  First, to the best of our knowledge, this is the first functional implementation of PCGRL in SMB, a platformer game with potentially infinite level length, that is arguably more complex than the GVGAI games studied in \cite{khalifa2020pcgrl}. Second, compared to tile-scale design in \cite{khalifa2020pcgrl}, the proposed approach is (required to be) faster as level segments are generated online through the latent vectors of pre-trained generators. Finally, we quantify Koster's \emph{fun} \cite{koster2013theory} as a function that maintains moderate levels of Kullback-Leibler divergence (KL-divergence) within a level teaching our RL agent to generate levels with such properties.

\section{Background}\label{sec:background}

A substantial body or literature has defined the area of PCG in recent years~\cite{togelius2011search,togelius2013procedural,shaker2016procedural,yannakakis2011experience,yannakakis2018artificial,summerville2018procedural,Kegel2020Puzzle,risi2020increasing,liu2021dlpcg}. In this section, we focus on related studies in PCG via reinforcement learning and approaches for online level generation.

\subsection{PCG via Reinforcement Learning}\label{sec:pcgrl}

Togelius \emph{et al.}~\cite{togelius2013procedural} have proposed three fundamental goals for PCG: ``\textit{multi-level multi-content PCG, PCG-based game design and generating complete games}''. To achieve these goals, various types of PCG methods and frameworks have been researched and applied in games since then~\cite{shaker2016procedural,togelius2011search,yannakakis2011experience,yannakakis2018artificial}. 
The development of machine learning (ML) has brought revolution to PCG~\cite{yannakakis2018artificial}. The combination of ML and PCG (PCGML)~\cite{summerville2018procedural} shows great potential compared with classical PCG methods; in particular, deep learning methods have been playing an increasingly important role in PCG in recent years~\cite{liu2021dlpcg}. Furthermore, PCG methods can be used to increase the generality in ML~\cite{risi2020increasing}. PCGML, however, is limited by the lack of training data that is often the case in games~\cite{summerville2018procedural}. Khalifa \emph{et al.}~\cite{khalifa2020pcgrl} proposed PCG via reinforcement learning (PCGRL) which frames level generation as a game and uses an RL agent to solve it. A core advantage of PCGRL compared with existing PCGML frameworks~\cite{summerville2018procedural} is that no training data is required. More recently, adversarial reinforcement learning has been applied to PCG~\cite{gisslen2021adversarial}; specifically, a PCGRL agent for generating different environments is co-evolved with a problem solving RL agent that acts in the generated environments~\cite{gisslen2021adversarial}. Engelsvoll \emph{et al.}~\cite{engelsvoll2020generating} applied a Deep Q-Network (DQN) agent to play SMB levels generated by a DQN-based level designer, which takes as input the latest columns of a played level and outputs new level columns in a tile-by-tile manner. Although, the work of \cite{engelsvoll2020generating} was the first attempt of implementing PCGRL in SMB, the emphasis in that work was not in the online generation of experience-driven PCG.

\subsection{Online Level Generation}\label{sec:onlinepcg}

Content generation that occurs in real-time (i.e. online) requires rapid generation times. Towards that aim, Greuter \emph{et al.}~\cite{greuter2003real} managed to generate ``pseudo infinite'' virtual cities in real-time via simple constructive methods. Johnson \emph{et al.}~\cite{johnson2010cellular} used cellular automata to generate infinite cave levels in real-time. In \cite{jennings2010polymorph}, the model named polymorph was proposed for dynamic difficulty adjustment during the generation of levels. Stammer \emph{et al.}~\cite{stammer2015player} generated personalised and difficulty adjusted levels in the 2D platformer \emph{Spelunky} (Mossmouth, LLC, 2008) while Shaker \emph{et al.} formed the basis of the experience-driven PCG framework \cite{yannakakis2011experience} by generating personalised platformer levels online \cite{shaker2010towards}. Shi and Chen~\cite{shi2016online} combined rule-based and learning-based methods to generate online level segments of high quality, called constructive primitives (CPs).
In the work of~\cite{shi2017learning}, a dynamic difficulty adjustment algorithm with Thompson Sampling~\cite{thompson1933likelihood} was proposed to combine these CPs.

\section{EDRL: Learn to Design Experiences via RL}\label{sec:framework}

\begin{figure*}[!tb]
    \centering
 \includegraphics[width=0.8\linewidth]{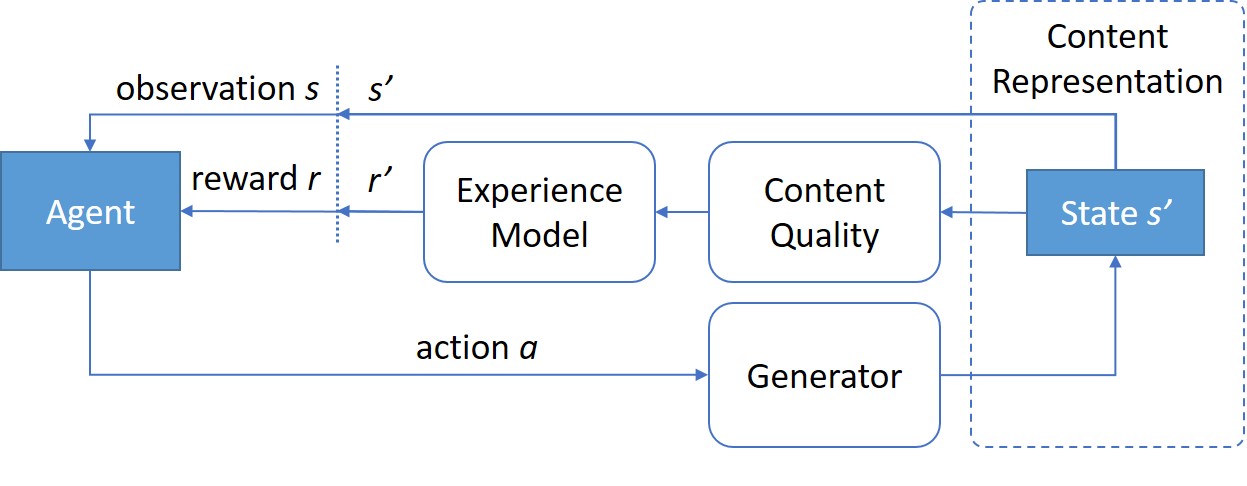}
    \caption{General overview of the EDRL framework interweaving elements of both the EDPCG and the PCGRL frameworks. White rounded boxes and blue boxes depict components of EDPCG \cite{yannakakis2011experience} and PCGRL \cite{khalifa2020pcgrl}, respectively. Content representation (i.e., environment in RL terms) is a common component across the two frameworks and is depicted with a dashed line.}
    \label{fig:EDRL}
\end{figure*}

Our aim is to introduce a framework that is able to generate endless, novel yet consistent and playable levels, ideally for any type of game. Given an appropriate level-segment generator, our level generator selects iteratively the suitable segment to be concatenated successively to the current segment to build a level. This process resembles a jigsaw puzzle game, of which the goal is to pick suitable puzzle pieces (i.e., level segment) and put them together to match a certain pattern or ``style''. 

Playing this jigsaw puzzle can be modelled as a Markov Decision Process (MDP)~\cite{bellman1957markovian}. In this MDP, a state $s$ models the current puzzle piece (level segment) and an action is the next puzzle piece to be placed. The reward $r(s,a)$ evaluates how these two segments fit. The next state $s'$ after selecting $a$ is set as $a$ itself assuming deterministic actions, i.e., $s'=MDP(s,a)=a$. An optimal action $a^*$ at state $s$ is defined as the segment that maximises the reward $r(s,a)$ if being placed after segment $s$, i.e., $a^*=\arg\max_{a\in \mathcal{A}}r(s,a)$. $\mathcal{A}$ denotes the actions space, i.e., the set of all possible segments. An RL agent is trained to learn the optimal policy $\pi^*$ that selects the optimal segments, thus $a^*=\pi^*(s)$.
Endless-level generation can be achieved if this jigsaw puzzle game is played infinitely.

One could argue that the jigsaw puzzle described above does not satisfy the Markov property as the level consistency and diversity should be measured based on the current and the concatenated segment. When a human plays such a game, however, she can only perceive the current game screen. In fast-paced reactive games, the player's short term memory is highly active during reactive play and, thus episodic memory of the game's surroundings is limited to a local area around play~\cite{tulving2002episodic}.
Therefore, we can assume that the Markov property is satisfied to a good degree within a suitable length of such level segments.

The general overview of the EDRL framework is shown in Fig.~\ref{fig:EDRL}. The framework builds on EDPCG \cite{yannakakis2011experience} and PCGRL \cite{khalifa2020pcgrl} and extends them both by enabling experience-driven PCG via the RL paradigm. According to EDRL an RL agent learns to design content with certain player experience and aesthetic aspects (\emph{experience model}) by interacting with the RL environment which is defined through a \emph{content representation}. The \emph{content quality} component guarantees that the experience model will consider content of certain quality (e.g., tested via gameplay simulations). The RL designer takes an action that corresponds to a generative act that alters the state of the represented content ($s'$) and receives a reward $r'$ through the \emph{experience model} function. The agent iteratively traverses the design (state-action representation) and experience (reward) space to find a design policy that optimises the \emph{experience model}. 

The framework is directly applicable to any game featuring levels that can be segmented and represented rapidly through a compressed representation such as a latent vector. This includes Atari-like 2D games \cite{bellemare2013arcade,perez2019general} but can also include more complex 3D games (e.g., VizDoom \cite{kempka2016vizdoom}) if latent vectors are available and can synthesise game content. 

\subsection{Novelty Of EDRL}\label{sec:diff}
The presented EDRL framework sits at the intersection of experience-driven PCG and PCGRL being able to create personalised (i.e., experience-tailored) levels via RL agents in a real-time manner. EDRL builds on the core principles of PCGRL \cite{khalifa2020pcgrl} but it extends it in a number of ways. First, vanilla PCGRL focuses on training an RL agent to design levels from scratch. Our framework, instead, teaches the agent to learn to select suitable level segments based on content and game-play features. Another key difference is the action space of RL agents. Instead of tile-scale design, our designer agent selects actions in the latent space of the generator and uses the output segment to design the level in an online manner. The length of the level is not predefined in our framework. Therefore, game levels can in principle be generated and played endlessly. 

For designing levels in real-time, we consider the diversity of new segments compared to the ones created already, which yields reward functions for fun and deviation over time; earlier work (e.g., \cite{shi2017learning,jennings2010polymorph}) largely focused on objective measures for dynamic difficulty adjustment. Additionally, we use a repairer that corrects level segments without human intervention, and game-playing agents that ensure their playability. It is a hard challenge to determine which level segment will contribute best to the level generation process---i.e. a type of credit assignment problem for level generation. To tackle this challenge, Shi and Chen~\cite{shi2017learning} formulated dynamic difficulty adjustment as an MDP~\cite{bellman1957markovian} with binary reward and a Thompson sampling method. We, instead, frame online level generation as an RL game bounded by any reward function which is not limited to difficulty, but rather to player experience. 

\section{EDRL for Mario: MarioPuzzle}\label{sec:mariopuzzle}

\begin{figure}[!tb]
    \centering
    \includegraphics[width=1\columnwidth]{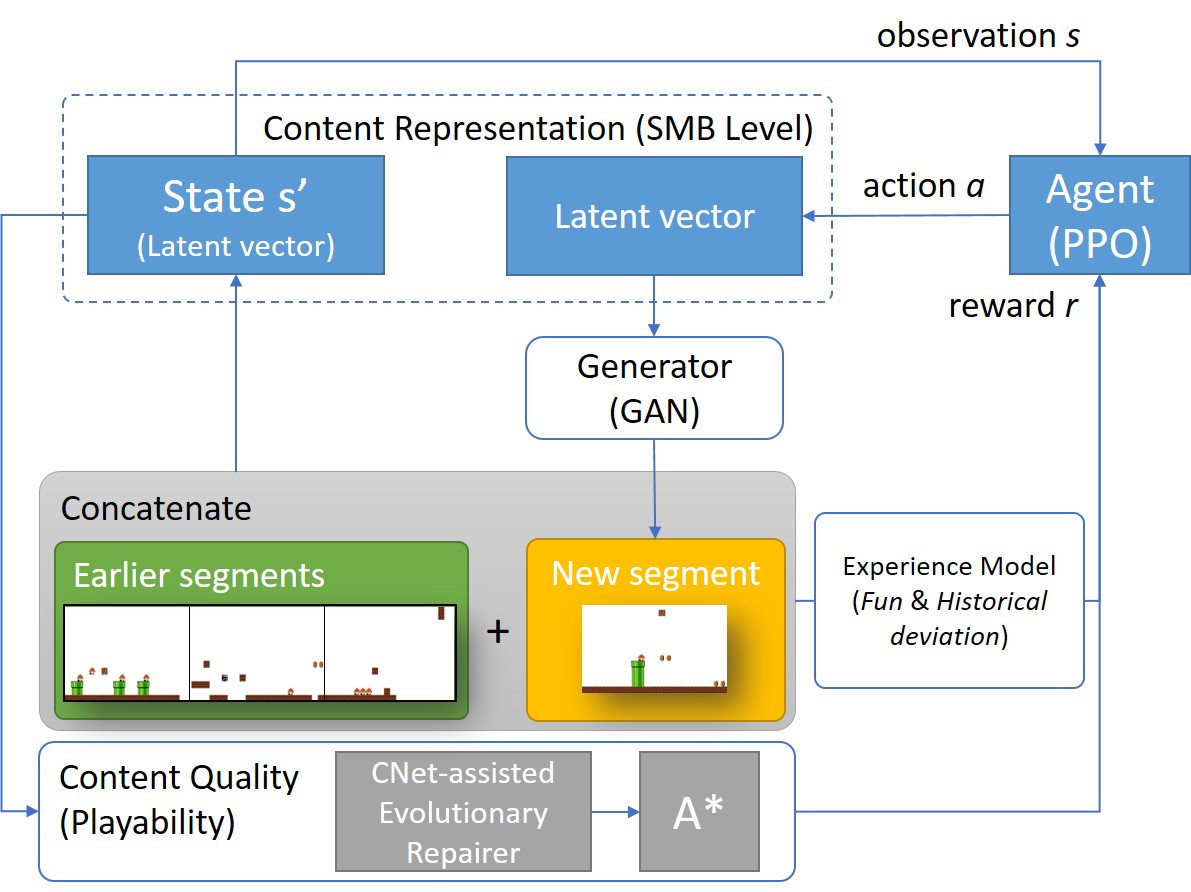}
    \caption{\label{fig:mariopuzzle}Implementing EDRL on Super Mario Bros: the \emph{MarioPuzzle} framework.}
\end{figure}

In this section we introduce an implementation of EDRL for SMB level generation and the specific reward functions designed and used in our experiments (Section \ref{sec:reward_design}). EDRL in this implementation enables \emph{online} and \emph{endless} generation of content under functional (i.e., playbility) and aesthetic (e.g., fun) metrics. As illustrated in Fig. \ref{fig:mariopuzzle}, the implementation for SMB---namely \emph{MarioPuzzle}\footnote{Available on GitHub: \url{https://github.com/SliverySky/mariopuzzle}}---features three main components: (i) a generator and repairer of non-defective segments, (ii) an artificial SMB player that tests the playability of the segments and (iii) an RL agent that plays this \emph{MarioPuzzle} endless-platform generation game. We describe each component in dedicated sections below. 

\begin{algorithm}[htbp]
\caption{\label{algo:mariopuzzle}Training procedure of \emph{MarioPuzzle}.}
\begin{algorithmic}[1]
\REQUIRE{$\mathcal{G}:[-1,1]^{32}\mapsto segment$, trained GAN}
\REQUIRE{$\mathcal{F}:segment \mapsto segment$, trained repairer}
\REQUIRE{$\mathcal{P}:segment \mapsto [0,1]$, $A^*$ agent}
\REQUIRE{$\pi:[-1,1]^{32} \mapsto [-1,1]^{32}$, RL agent}
\REQUIRE{$Reward$, reward function}
\REQUIRE{$T$: maximum number of training epochs}
\REQUIRE{$N$: maximum segment number of one game}
\STATE{$t \leftarrow 0$}
\WHILE{$t<T$}
\STATE{$PreS\leftarrow Empty\ List$}
\STATE{$n \leftarrow 0$}
\STATE{$s_n\leftarrow NULL$} 
\REPEAT
\STATE{$s\leftarrow$ uniformly sampled from $[-1,1]^{32}$}
\STATE{$S\leftarrow \mathcal{G}(s)$} \COMMENT{Generate a segment} 
\STATE{$S\leftarrow \mathcal{F}(S)$} \COMMENT{Repair the segment if applicable}
\STATE{$isPlayable\leftarrow \mathcal{P}(S)$} \COMMENT{Play this segment}
\IF{$isPlayable$}
\STATE{$s_n\leftarrow v$} 
\ENDIF
\UNTIL{$s_n$ is not $NULL$} 
\STATE{Add $S$ to $PreS$}
\WHILE{$isPlayable$ and $n<N$} 
\STATE{$a_n \leftarrow \pi(s_n)$} \COMMENT{Select an action}
\STATE{$S\leftarrow \mathcal{G}(a_n)$} \COMMENT{Generate a segment} 
\STATE{$S\leftarrow \mathcal{F}(S)$} \COMMENT{Repair the segment if applicable}
\STATE{$isPlayable\leftarrow \mathcal{P}(S)$} \COMMENT{Play this segment}
\STATE{$score_{s_n,a_n} \leftarrow Reward(S)$ with previous $PreS$} \COMMENT{According to metrics}
\STATE{Update $\pi$ with $score_{s_n,a_n}$} 
\STATE{Update $PreS$ with $S$} \COMMENT{According to metrics}
\STATE{$n=n+1$}
\ENDWHILE
\STATE{$t=t+1$}
\ENDWHILE
\RETURN{$\pi$}
\end{algorithmic}
\end{algorithm}

\subsection{Generate and Repair via the Latent Vector}

As presented earlier in Section \ref{sec:framework}, both the actions and states in this MDP represent different level segments. Our framework naturally requires a level segment generator to operate. For that purpose we use and combine the \emph{MarioGAN}~\cite{volz2018evolving} generator and \emph{CNet-assisted Evolutionary Repairer}~\cite{shu2020cnet} to, respectively, generate and repair the generated level segments. The CNet-assisted Evolutionary Repairer has shown to be capable of determining wrong tiles in segments generated by MarioGAN~\cite{volz2018evolving} and repairing them~\cite{shu2020cnet}; hence, the \emph{correctness} of generated segments is guaranteed. In particular, we train a GAN\footnote{\url{https://github.com/schrum2/GameGAN}} on fifteen SMB levels of three types (overworld, underground and athletic) in VGLC~\cite{summerville2016vglc}. The CNet-assisted Evolutionary Repairer trained by \cite{shu2020cnet} is used directly and unmodified \footnote{\url{https://github.com/SUSTechGameAI/MarioLevelRepairer}}.

It is important to note that we are not using a direct tile-based representation as in \cite{khalifa2020pcgrl} or one-hot encoded tiles as in \cite{engelsvoll2020generating}; instead we use the latent vector of \emph{MarioGAN} to represent the agent's actions and states for the RL agent. Thus, the selected action $a$ or state $s$ are sampled from the latent space, rather than the game space.

\subsection{The AI Player}

The $A^*$ agent in the Mario AI framework~\cite{shaker20112010} is used as an artificial player for determining the playability of generated segments\footnote{\url{https://github.com/amidos2006/Mario-AI-Framework}}. 

\subsection{The RL Designer}

 \emph{MarioPuzzle} is embedded into the OpenAI gym~\cite{brockman2016openai} and the PPO algorithm\cite{schulman2017proximal} is used to train our agent~\cite{pytorchrl}. The training procedure of \emph{MarioPuzzle} is described in Algorithm \ref{algo:mariopuzzle}. The actions and states are represented by latent vectors of length $32$. 
When training the PPO agent, the initial state is a playable segment randomly sampled from the latent space of the trained segment generator, as shown in Algorithm \ref{algo:mariopuzzle}. The latent vector of the current level segment feeds the current observation of the PPO agent. Then, an action (i.e., a latent vector) is selected by the PPO agent and used as an input of the generator to generate a new segment. The repairer determines and fixes broken pipes in the new segment. The fixed segment is concatenated to the earlier one and the $A^*$ agent tests if the addition of the new segment is playable. Then, the latent vector of this new segment (same as action) is returned as a new observation to the agent. The agent receives an immediate reward for taking an action in a particular state. The various reward functions we considered in this study are based on both content and game-play features and are detailed below. 

\subsection{Reward Functions}
\label{sec:reward_design}

Designing a suitable \emph{reward} function for the RL agent to generate desired levels is crucial. In this implementation of EDRL, we design three metrics that formulate various reward functions ($Reward$ in Algorithm \ref{algo:mariopuzzle}) aiming at guiding the RL agent to learn to generate playable levels with desired player experiences.

\subsubsection{Moderating Diversity Makes Fun!}\label{sec:variance}


Koster's \emph{theory of fun} \cite{koster2013theory} suggests that a game is fun when the patterns a player perceives are neither too unfamiliar (i.e., changeling) nor too familiar (i.e., boring). Inspired by this principle, when our agent concatenates two segments, we assume it should keep the diversity between them at moderate levels; too high diversity leads to odd connections (e.g., mix of styles) whereas too low diversity yields segments that look the same. To do so, we first define a diversity measure, and then we moderate diversity as determined from human-designed levels.

When a player plays through a level, the upcoming level segment is compared with previous ones for diversity. We thus define the diversity of a segment through its dissimilarity to previous segments. While there have been several studies focusing on quantifying content similarity (e.g., \cite{isaksen2017semantic}), we adopt the tile-based KL-divergence~\cite{lucas2019tile} as a simple and efficient measure of similarity between segments. More specifically diversity, $D$, is formulated as follows: 
\begin{equation}
    D(S) = \frac{1}{n+1}\sum_{i=0}^{n} KL(S,S_{W_i}),
    \label{eq:variance}
\end{equation}  
\noindent where $S$ is a generated level segment with height $h$ and width $w$; $KL(a,b)$ is the tile-based KL-divergence that considers the standard KL-Divergence between the distributions over occurrences of tile patterns in two given level segments, $a$ and $b$~\cite{lucas2019tile}. We define a window $W$ with the same size as $S$. $S_{W}$ represents the segment contained in the window $W$. A sliding window moves from the position of $S$ ($S_{W_0}=S$) to the previous segment $n$ times with stride $d$. The parameter $n$ limits the number of times that the window moves. According to \eqref{eq:variance}, larger $n$ values consider more level segments in the past. After preliminary hyper-parameter tuning, we use a $2 \times 2$ window and $\epsilon=0.001$ for calculating tile-based KL-divergence~\cite{lucas2019tile}. $n$ and $d$ are set as $3$ and $7$ in this work.

Once we have diversity defined, we attempt to moderate diversity by considering the fifteen human-authored SMB levels used for training our GAN. We thus calculate the average and standard deviation of $D$ for all segments across the three different level types of our dataset. Unsurprisingly, Table \ref{tab:variance} shows that different level types yield significantly different degrees of diversity. It thus appears that by varying the degree of diversity we could potentially vary the design style of our RL agent. Randomly concatenating level segments, however, cannot guarantee moderated diversity as its value depends highly on the expressive range of the generator. On that end, we moderate the diversity of each segment in predetermined ranges---thereby defining our \emph{fun} ($F$) reward function---as follows.

\begin{equation}\label{eq:fun}
     F(S) =\begin{cases}
    -(D(S)-u)^2, & \text{ if } D(S) > u\\
    -(D(S)-l)^2, & \text{ if } D(S) < l\\
    0, & \text{ otherwise},\\
  \end{cases}
  \end{equation}
\noindent where $l$ and $u$ denote the lower and upper bounds of diversity, respectively. According to the analysis on human-authored SMB levels (Table \ref{tab:variance}), we assume that the diversity of \emph{overworld} levels is moderate and arbitrarily set $l=0.60-0.34=0.26$ and $u=0.60+0.34=0.94$, based on the diversity values obtained for this type of levels. 

\begin{table}[!tb]
\centering
\caption{\label{tab:variance}Average diversity values and corresponding standard deviations of segments across the three types of SMB levels.}
\begin{tabular}{c|c|r|c}
\toprule
\multicolumn{1}{c|}{Type} & \multicolumn{1}{c|}{\#Level} & \multicolumn{1}{c|}{$\sum$\#Segment} & $D$ [Eq. \eqref{eq:variance}]\\
\midrule
Overworld & 9 & $1,784$ & $0.60\pm0.34$ \\
Athletic & 4 & $534$ &  $0.32\pm0.25$\\
Underground &  2 & $305$ & $1.11\pm0.59$\\
\midrule
Total& 15 & $2,623$ & $0.60\pm0.42$ \\
\bottomrule
\end{tabular}
\end{table}

\subsubsection{Historical Deviation}

Inspired by the notion of novelty score \cite{lehman2008exploiting}, we consider \emph{historical deviation}, $H$, as a measure that encourages our agent to generate segments that deviate from earlier creations. In particular, we define $H$ of a segment as the average similarity of the $k$ most similar segments among the $m>k$ previous segments, as formalised in \eqref{eq:novelty}. 

\begin{equation}
    H(S)=\frac{1}{k}\sum_{j=1}^{k}KL(S, S_{I_{j}}),
    \label{eq:novelty}
\end{equation}
\noindent where $I_{j}$ is the index of the ${j}^{th}$ most similar segment among previous $m$ segments compared to $S$; $m$ represents the number of segments that our RL agent holds in its memory. This parameter is conceptually similar to the \emph{sparseness} parameter in novelty search~\cite{lehman2011novelty}. After preliminary hyper-parameter tuning, $m$ and $k$ are set as $20$ and $10$, respectively, in the experiments of this paper.

\subsubsection{Playablility}\label{sec:pmetric}

The playablility of a newly generated segment is tested by an $A^*$ agent. The generated segment, however, is not played on its own; its concatenation to the three previous segments is played as a whole instead, to ensure that the concatenation will not yield unplayable levels. When testing, Mario starts playing from its ending position in the previous level segment. 
A segment $S$ is determined as playable only if Mario succeeds in reaching the right-most end of $S$.

Playability is essential in online level generation. Therefore, when an unplayable segment is generated, the current episode of game design ends---see Algorithm \ref{algo:mariopuzzle}---and the reward returned, $P(S)$, is set as the number of segments that the $A^*$ Mario managed to complete.
With such a reward function the agent is encouraged to generate the longest playable levels possible. 

\section{Design Experiments Across Reward Functions}\label{sec:effective}

In this section, we test the effectiveness of the metrics detailed in Section \ref{sec:reward_design} through empirical experimentation. We use each of the fun and historical deviation metrics as independent reward functions and observe their impact on generated segments. Then, we use combinations of these metrics and playability to form new reward functions.

\subsection{Experimental Details and Results}

\def\old{
\begin{table*}[!tb]
    \centering    
    \caption{\label{tab:evaluate}
    Evaluation metrics of levels generated across different RL agents over 300 levels each. The Table presents average values and corresponding standard deviations across the 300 levels. $\pi_*$ refers to the trained policy using the $*$ reward function. $\pi_R$ refers to a random agent that designs SMB levels. $\overline{F}$, $\overline{H}$ and $\overline{P}$ are respectively the $F$, $H$ and $P$ values averaged over playable segments by the $A^*$ agent; in addition to $\overline{F}$, the $\overline{F}_b$ value appearing in square brackets refers to the average percentage of playable segments within the bounds of moderate diversity (cf. Section \ref{sec:variance}). In addition to the three evaluation metrics, the Table presents the number of gaps, pipes, enemies, bullet blocks, coins, and question-mark blocks that exist in the generated playable segments. Values in bold indicate the highest value in the column.} 
    
    
    \setlength{\tabcolsep}{3pt}
    \begin{tabular}{c||c|c|c||c|c|c|c|c|c}
    \toprule
    \multirow{2}{*}{Agent} & \multicolumn{3}{c||}{Evaluation metrics} & \multicolumn{6}{c}{Number of level elements in generated segments} \\  
      & $\overline{F}$ [$\overline{F}_b$] & $\overline{H}$ & $\overline{P}$ &Gaps&Pipes&Enemies&Bullets&Coins&Question-marks \\
     \midrule
$\pi_{F}$  & \textbf{-0.005$\pm$0.044} [\textbf{87.1$\pm$14.1}] & 0.91$\pm$0.22 & 29.6$\pm$28.3 &0.60$\pm$0.40 & 0.43$\pm$0.17 & \textbf{2.11$\pm$0.63} & 0.05$\pm$0.13 & 0.64$\pm$0.52 & \textbf{1.00$\pm$0.59}\\   
$\pi_{H}$  & -0.092$\pm$0.092 [57.0$\pm$18.8] & \textbf{1.60$\pm$0.29} & 24.2$\pm$21.8 &0.73$\pm$0.34 & 0.40$\pm$0.31 & 1.48$\pm$0.58 & 0.09$\pm$0.10 & 1.10$\pm$0.62 & 0.68$\pm$0.55\\  
$\pi_{FH}$ & -0.065$\pm$0.086 [63.4$\pm$21.8] & 1.47$\pm$0.35 & 16.4$\pm$16.2&0.74$\pm$0.37 & \textbf{0.49$\pm$0.33} & 1.69$\pm$0.74 & 0.11$\pm$0.19 & 1.06$\pm$0.79 & 0.53$\pm$0.59\\
\midrule
$\pi_{P}$ & -0.023$\pm$0.013 [76.7$\pm$5.1] & 1.04$\pm$0.07 & 97.3$\pm$11.8
&0.12$\pm$0.04 & 0.15$\pm$0.04 & 1.58$\pm$0.15 & 0.10$\pm$0.03 & 2.37$\pm$0.25 & 0.29$\pm$0.09
\\
$\pi_{FP}$& -0.032$\pm$0.017 [75.2$\pm$5.2] & 1.17$\pm$0.07 & 96.6$\pm$14.4 &0.18$\pm$0.05 & 0.17$\pm$0.04 & 1.28$\pm$0.17 & 0.10$\pm$0.04 & \textbf{2.51$\pm$0.25} & 0.46$\pm$0.11 \\  

$\pi_{HP}$ & -0.037$\pm$0.020 [74.2$\pm$5.7] & 1.18$\pm$0.08 & 97.0$\pm$13.9  &0.18$\pm$0.07 & 0.22$\pm$0.05 & 1.40$\pm$0.20 & 0.10$\pm$0.04 & 2.48$\pm$0.29 & 0.46$\pm$0.18\\   
$\pi_{FHP}$ & -0.034$\pm$0.018 [74.0$\pm$6.0] & 1.18$\pm$0.09 & \textbf{97.4$\pm$13.7}  &0.23$\pm$0.17 & 0.18$\pm$0.09 & 1.19$\pm$0.40 & \textbf{0.12$\pm$0.04} & 2.43$\pm$0.36 & 0.55$\pm$0.13\\  
\midrule
   $\pi_{R}$ & -0.064$\pm$0.098 [64.3$\pm$21.9] & 1.43$\pm$0.34 & 16.0$\pm$14.3 &\textbf{0.82$\pm$0.44} & 0.45$\pm$0.36 & 1.53$\pm$0.75 & 0.10$\pm$0.18 & 0.94$\pm$0.71 & 0.52$\pm$0.55\\  
   \bottomrule
    \end{tabular}
\end{table*}
}

\begin{table*}[!tb]
    \centering    
    \caption{\label{tab:evaluate}
    Evaluation metrics of levels generated across different RL agents over 300 levels each. The Table presents average values and corresponding standard deviations across the 300 levels. $\pi_*$ refers to the trained policy using the $*$ reward function. $\pi_R$ refers to a random agent that designs SMB levels. $\overline{F}$, $\overline{H}$ and $\overline{P}$ are respectively the $F$, $H$ and $P$ values averaged over playable segments by the $A^*$ agent; in addition to $\overline{F}$, the $\overline{F}_b$ value appearing in square brackets refers to the average percentage of playable segments within the bounds of moderate diversity (cf. Section \ref{sec:variance}). In addition to the three evaluation metrics, the Table presents the number of gaps, pipes, enemies, bullet blocks, coins, and question-mark blocks that exist in the generated playable segments. Values in bold indicate the highest value in the column.} 
    
    
    \setlength{\tabcolsep}{3pt}
    \begin{tabular}{c||c|c|c||c|c|c|c|c|c}
    \toprule
    \multirow{2}{*}{Agent} & \multicolumn{3}{c||}{Evaluation metrics} & \multicolumn{6}{c}{Number of level elements in generated segments} \\  
      & $\overline{F}$ [$\overline{F}_b$] & $\overline{H}$ & $\overline{P}$ &Gaps&Pipes&Enemies&Bullets&Coins&Question-marks \\
     \midrule
$\pi_{F}$  & \textbf{-0.005$\pm$0.044} [\textbf{87.1$\pm$14.1}] &  0.86$\pm$0.28 & 29.6$\pm$28.3 &0.60$\pm$0.40 & 0.43$\pm$0.17 & \textbf{2.11$\pm$0.63} & 0.05$\pm$0.13 & 0.64$\pm$0.52 & \textbf{1.00$\pm$0.59}\\   
$\pi_{H}$  & -0.092$\pm$0.092 [57.0$\pm$18.8] & \textbf{1.43$\pm$0.32}  & 24.2$\pm$21.8 &0.73$\pm$0.34 & 0.40$\pm$0.31 & 1.48$\pm$0.58 & 0.09$\pm$0.10 & 1.10$\pm$0.62 & 0.68$\pm$0.55\\  
$\pi_{FH}$ & -0.065$\pm$0.086 [63.4$\pm$21.8] & 1.38$\pm$0.38  & 16.4$\pm$16.2&0.74$\pm$0.37 & \textbf{0.49$\pm$0.33} & 1.69$\pm$0.74 & 0.11$\pm$0.19 & 1.06$\pm$0.79 & 0.53$\pm$0.59\\
\midrule
$\pi_{P}$ & -0.023$\pm$0.013 [76.7$\pm$5.1] & 0.72$\pm$0.07 & 97.3$\pm$11.8
&0.12$\pm$0.04 & 0.15$\pm$0.04 & 1.58$\pm$0.15 & 0.10$\pm$0.03 & 2.37$\pm$0.25 & 0.29$\pm$0.09
\\
$\pi_{FP}$& -0.032$\pm$0.017 [75.2$\pm$5.2] &  0.83$\pm$0.09  & 96.6$\pm$14.4 &0.18$\pm$0.05 & 0.17$\pm$0.04 & 1.28$\pm$0.17 & 0.10$\pm$0.04 & \textbf{2.51$\pm$0.25} & 0.46$\pm$0.11 \\  

$\pi_{HP}$ & -0.037$\pm$0.020 [74.2$\pm$5.7] & 0.84$\pm$0.09  & 97.0$\pm$13.9  &0.18$\pm$0.07 & 0.22$\pm$0.05 & 1.40$\pm$0.20 & 0.10$\pm$0.04 & 2.48$\pm$0.29 & 0.46$\pm$0.18\\   
$\pi_{FHP}$ & -0.034$\pm$0.018 [74.0$\pm$6.0] & 0.84$\pm$0.09  & \textbf{97.4$\pm$13.7}  &0.23$\pm$0.17 & 0.18$\pm$0.09 & 1.19$\pm$0.40 & \textbf{0.12$\pm$0.04} & 2.43$\pm$0.36 & 0.55$\pm$0.13\\  
\midrule
   $\pi_{R}$ & -0.064$\pm$0.098 [64.3$\pm$21.9] & 1.35$\pm$0.37  & 16.0$\pm$14.3 &\textbf{0.82$\pm$0.44} & 0.45$\pm$0.36 & 1.53$\pm$0.75 & 0.10$\pm$0.18 & 0.94$\pm$0.71 & 0.52$\pm$0.55\\  
   \bottomrule
    \end{tabular}
\end{table*}

When calculating $F$, the size of the sliding window and the segment are both $14\times 14$ and the control parameters $n$ and $d$ of Eq. \eqref{eq:variance} are set as $3$ and $7$, respectively. When calculating $H$, the control parameters $m$ and $k$ of Eq. \eqref{eq:novelty} are set as $20$ and $10$, respectively. For each RL agent presented in this paper, a PPO algorithm is used and trained for $10^6$ epochs. 

In all results and illustrations presented in this paper, $F$, $H$ and $P$ refer to independent metrics that consider, respectively, the fun \eqref{eq:fun}, historical deviation \eqref{eq:novelty} and playability (cf. Section \ref{sec:pmetric}) of a level. The notation $\pi_{F}$ refers to the PPO agent trained solely through the $F$ reward function. Similarly, $\pi_{FH}$ refers to the agent trained with the sum of $F$ and $H$ as a reward, while $\pi_{FHP}$ refers to the one trained with the sum of $F$, $H$ and $P$ as a reward. When a reward function is composed by multiple metrics, each of the metrics included is normalised within $[0,1]$ based on the range determined by the maximum and minimum of its $1,000$ most recent values. 

For testing the trained agents, $30$ different level segments are used as initial states. Given an initial state, an agent is tested independently $10$ times for a maximum of $100$ segments---or until an unplayable segment is reached if $P$ is a component of the reward function. As a result, each agent designs $300$ levels of potentially different number of segments when $P$ is considered; otherwise, the generation of an unplayable segment will not terminate the design process, thus $100$ segments will be generated. For comparison, a random agent, referred to as $\pi_{R}$, is used to generate levels by randomly sampling up to $100$ segments from the latent space or till an unplayable one is generated.

\begin{figure}[!tb]
    \subfloat[\label{fig:F}$\pi_{F}$: surprisingly uninteresting levels when designing sorely for fun]{
    \includegraphics[width=1\linewidth]{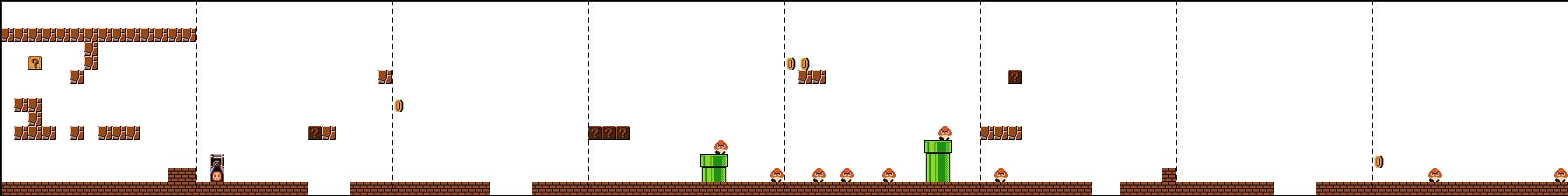}}
    \hfill
    \subfloat[\label{fig:H}$\pi_H$: highly diverse levels when designing historical deviation]{
    \includegraphics[width=1\linewidth]{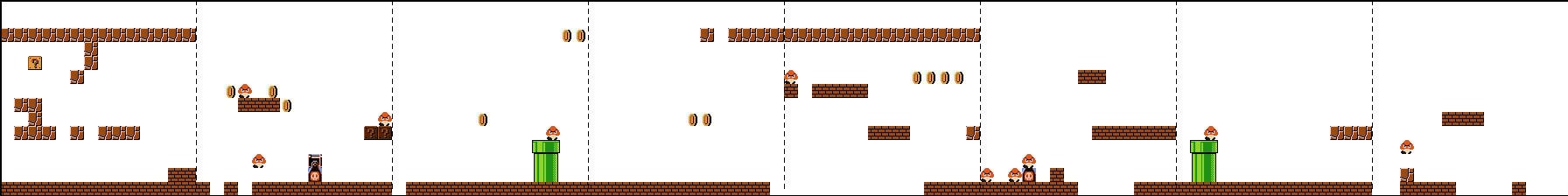}}
    \hfill
    \subfloat[\label{fig:FH}$\pi_{FH}$: levels balancing between fun and historical deviation]{    \includegraphics[width=1\linewidth]{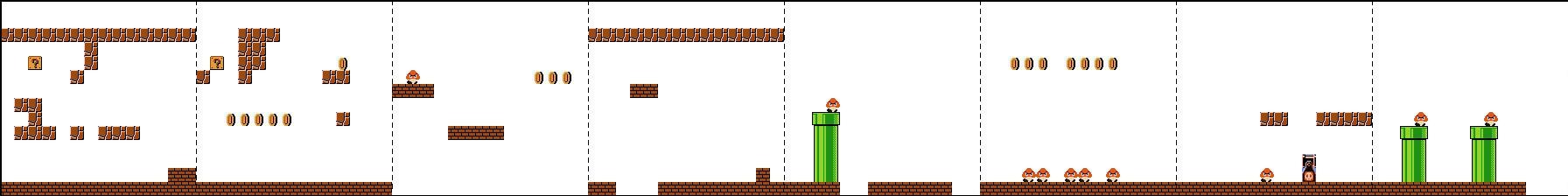}}
    \hfill
    \subfloat[\label{fig:P}$\pi_{P}$: levels with more ground tiles in assistance of $A^*$]{
    \includegraphics[width=1\linewidth]{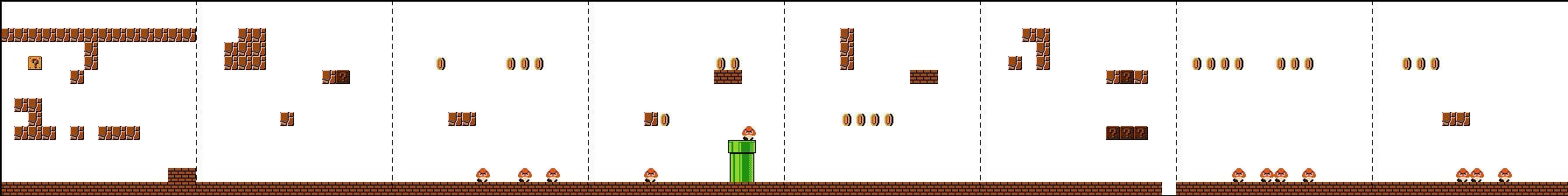}
    }
    \hfill
    \subfloat[\label{fig:FP}$\pi_{FP}$: playable levels with clearly repeated patterns ]{\includegraphics[width=1\linewidth]{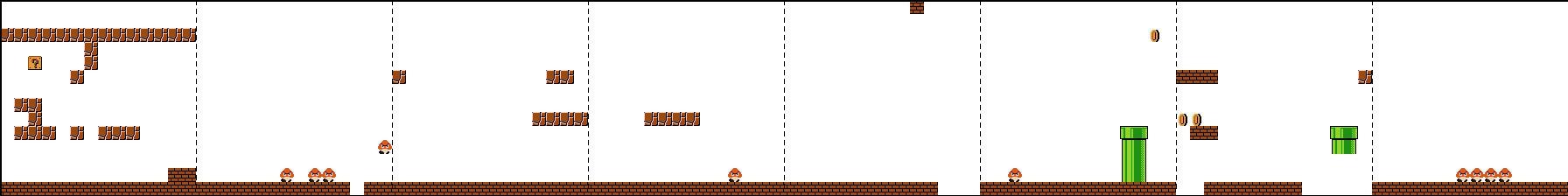}}
    \hfill
    \subfloat[\label{fig:HP}$\pi_{HP}$: playable and diverse levels with limited gaps]{    \includegraphics[width=1\linewidth]{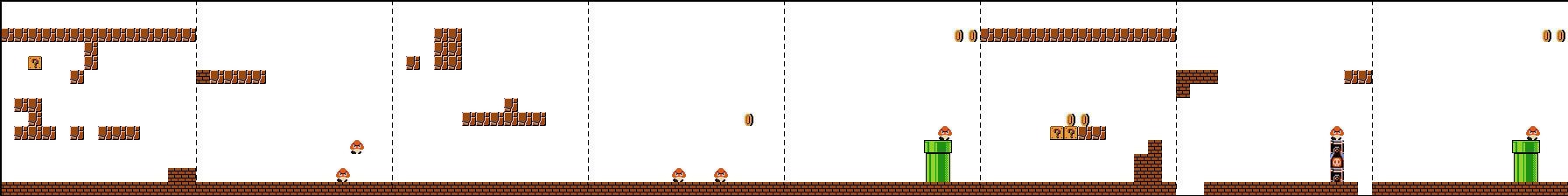}}
    \hfill
    \subfloat[\label{fig:FHP}$\pi_{FHP}$: fun, diverse and playable levels]{
        \includegraphics[width=1\linewidth]{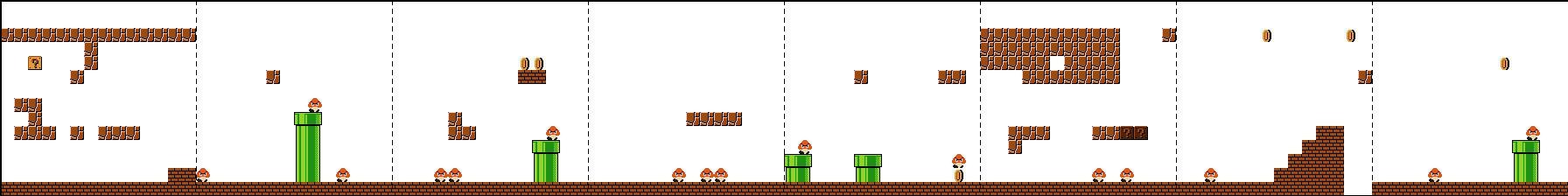}}
    \caption{\label{fig:segment_level}Example segments clipped from levels generated by RL agents trained with different reward functions. The key characteristics of each EDRL policy are outlined in the corresponding captions.}
\end{figure}

The degrees of fun, historical deviation, and playability of these generated levels are evaluated and summarised in Table~\ref{tab:evaluate}. Table \ref{tab:evaluate} also showcases average numbers of core level elements in generated levels, including pipes, enemies and question-mark blocks. Figure~\ref{fig:segment_level} illustrates a number of arbitrarily chosen segments clipped from levels generated by each RL agent. 

\subsection{Analysis of Findings}

The key findings presented in Table \ref{tab:evaluate} suggest that the evaluation functions of $\overline{F}$ and $\overline{H}$ across all generated levels reach their highest value when the corresponding reward functions are used. In particular, the $\pi_F$ policy yields appropriate levels of diversity in $87.1$ out of 100 segments, on average; higher than any other policy (see $\overline{F}_b$ values in Table~\ref{tab:evaluate}). Similarly $\overline{H}$ reaches its highest value on average ($1.43$) when policy $\pi_H$ is used. Finally, playabilty is boosted when $P$ is used as a reward function but it reaches its peak value of $97.4$ out of 100 segments on average when it is combined with both $F$ and $H$.

$\pi_F$ focuses primarily on enemy and question-mark block placement, but the levels it generates are far from looking diverse (see Fig. \ref{fig:F}). $\pi_H$ does not yield any distinct level element and results in rather diverse yet unplayable levels (see Fig. \ref{fig:H}). The combination of $F$ and $H$ without playability ($\pi_{FH}$) augments the number of pipes existent in the level and offers interesting but largely unplayable levels (see Table~\ref{tab:evaluate} and Fig. \ref{fig:FH}). 

\begin{table*}[!tb]
    \centering
    \caption{\label{tab:online}Generating 100-segment long levels. All values---expect from the number of failed generations---are averaged across 300 levels (10 trails each from 30 different initial level segments). In the ``Generation Time`` column, ``Sample'' refers to the time averaged over the total number of generated segments, including unplayable and playable ones, while ``Segment'' refers to the time averaged over successful level generations.}
    \begin{tabular}{c|c|c|c|c|c|c|c|c|c}
    \toprule
      Column index  & 1 &2&3&\multicolumn{2}{c|}{4}&\multicolumn{2}{c|}{5}&\multicolumn{2}{c}{6}\\
\hline
    \multirow{2}{*}{Agent}     & \multirow{2}{*}{Resampling method}             & \multirow{2}{*}{Failed generations} &\multirow{2}{*}{Unplayable segments}& \multicolumn{2}{c|}{Resamples}&\multicolumn{2}{c|}{Generation Time (s)} & \multicolumn{2}{c}{Faulty tiles}\\
    & & & & Max & Total & Segment & Sample & Original & Repaired\\ \midrule
        \multirow{2}{*}{$\pi_{FH}$}  &   Random & 205/300 & 6.15 & 1.97 &7.24 &1.09& 1.02 &50.2&8.4\\
     & Policy  & 196/300 & 6.36 & 2.18 &7.61 &1.11& 1.03 &54.1&10.2\\
     \midrule
    \multirow{2}{*}{$\pi_{FHP}$} & Random &0/300 & 0.10 & 0.11 &0.11 &0.79& 0.79 &6.1&0.3 \\ 
     &  Policy&5/300 & 0.12 & 0.11 &0.12 &0.79& 0.79 &6.1&0.3\\ 
    \bottomrule
    \end{tabular}
    \end{table*}

\begin{figure*}[!tb]
    \centering
    \includegraphics[width=1\linewidth]{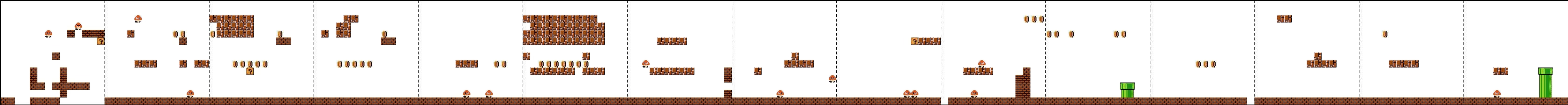}\\
    \vspace{.2em}
    \includegraphics[width=1\linewidth]{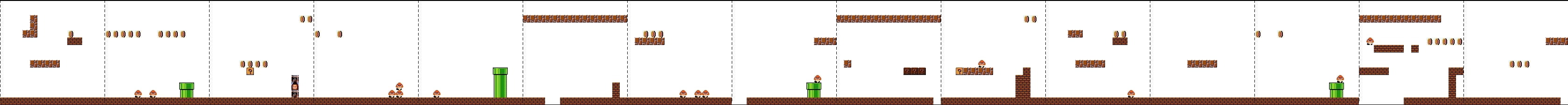}
    \caption{\label{fig:test1}Example of partial game levels generated online by the $\pi_{FHP}$ agent.}
\end{figure*}

When $P$ is considered by the reward function, the outcome is a series of playable levels (i.e., the average $P$ value is over 96 out of 100 segments per level). The most interesting agent from all the combinations of $P$ with the other two reward functions, however, seems to be the policy that is trained on all three ($\pi_{FHP}$; see Fig. \ref{fig:FHP}). That policy yields highly playable levels that maintain high levels of fun---i.e. on average, 74 out of 100 segments reach appropriate levels of diversity---and historical deviation ($0.84$ on average). 
The $\pi_{FHP}$ agent appears to design more playable levels, but at the same time, the generated levels yield interesting patterns and have more gaps, coins and bullet blocks that increase the challenge and augment the exploration paths available for the player (see Table~\ref{tab:evaluate} and Fig.~\ref{fig:FHP}). Based on the performance of $\pi_{FHP}$---compared against all other SMB level design agents explored---the next section puts this agent under the magnifying glass and analyses its online generation capacities.

\section{Online Endless-level Generation}\label{sec:onlinetest}

To test the performance of the trained $\pi_{FHP}$ agent for online level generation, we let it generate 300 levels (10 trials from 30 initial segments) composed of 100 playable segments each; although, in principle, the agent can generate endless levels online. The results obtained are shown in Table~\ref{tab:online} and are discussed in this section. As a baseline we compare the performance of $\pi_{FHP}$ against $\pi_{FH}$, i.e., an agent that does not consider playability during training.

When generating a level, it is necessary to repair the defects and test its playability. The faulty tiles of the generated segments before and after repairing are detected by CNet~\cite{shu2020cnet} (see column 6 in Table~\ref{tab:online}). Clearly, the levels generated by policy $\pi_{FH}$ feature more faulty tiles than levels generated by $\pi_{FHP}$. After visualising the generated levels, we observe that the pipes in the levels generated by $\pi_{FH}$ have diverse locations. Most of the pipes in levels generated by $\pi_{FHP}$, instead, are located in similar positions (i.e., near the middle bottom of the segment). We assume that the segments with this pattern are more likely to be chosen by the $\pi_{FHP}$ agent as they are easier to complete by the testing agent. 

The repairer operation by the CNet-assisted Evolutionary Algorithm only satisfies the logical constraints but does not guarantee the playability of the level. If a generated segment is unplayable, the RL agent will re-sample a new segment with one of the two methods: either by (i) sampling a new action according to its policy (ii) sampling a new action randomly from a normal distribution, and then clipping it into $[-1,1]^{32}$. This resampling process repeats until a playable segment is obtained or a sampling counter reaches $20$. Note that this resampling technique is only used in this online level generation test to ensure the generation of levels composed by playable segments only. Resampling is not enabled during training as an RL agent trained with playability as part of its reward is expected to learn to generate playable segments.

To evaluate the real-time generation efficiency of \emph{MarioPuzzle} we record the number of times resampling is required (see column 3 in Table \ref{tab:online}), the total number of resamplings (see column 4 in Table \ref{tab:online}), and the time taken for generating a level of 100 playable segments (see column 5 in Table \ref{tab:online}). 
Unsurprisingly, the $\pi_{FH}$ agent---which is trained without playability as part of its reward---appears to generate more unplayable segments and is required to resample more.

As a baseline of real-time performance we asked three human players (students in our research group) to play 8 levels that are randomly selected from the level creations of $\pi_{FHP}$. Their average playing time for one segment was $2.7s$, $3.1s$ and $7.6s$. It thus appears that the average segment generation time (see column 5 in Table~\ref{tab:online}) of $\pi_{FHP}$ is acceptable for online generation when compared to the time needed for a human player to complete one segment. 

According to Table~\ref{tab:online}, $\pi_{FHP}$ with random resampling never fails in generating playable 100-segment long levels. Comparing $\pi_{FHP}$ with $\pi_{FH}$, it is obvious that integrating playability into the reward function can reduce the probability of generating unplayable segments and resampling times, and, in turn, make the online generation of playable levels easier and faster. 

Figure \ref{fig:test1} displays examples from the levels generated by $\pi_{FHP}$. The EDRL designer resolves the playability issues by placing more ground tiles while maintaining appropriate levels of $F$ and $H$. It is important to remember that EDRL in this work operates and designs fun, diverse and playable levels for an $A^*$ playing agent. The outcome of Fig. \ref{fig:test1} reflects on the attempt of the algorithm to maximise all three reward functions for this particular and rather predictable player, thereby offering largely linear levels without dead-ends, with limited gaps (ensuring playability) and limited degrees of exploration for the player. The resulting levels for agents that depict more varied gameplay are expected to vary substantially.

\section{Discussion and Future Work}\label{sec:discussion}

In this paper we introduced EDRL as a framework, and instantiated it in SMB as \emph{MarioPuzzle}. We observed the capacity of \emph{MarioPuzzle} to generate endless SMB levels that are playable and attempt to maximise certain experience metrics. As this is the first instance of EDRL in a game, there is a number of limitations that need to be explored in future studies; we discuss these limitations in this section.

By integrating playability in the reward function we aimed to generate playable levels that are endless in principle~\cite{togelius20102009}. As a result, the RL agent generates more ground tiles to make it easier for the $A^*$ to pass though. The generated levels highly depend both on the reward function and the behaviour of the test agent. Various human-like agents and reward functions will need to be studied in future work to encourage the RL agent to learn to select suitable level segments for different types of players. Moreover, the behaviour of human-like agents when playing earlier segments can be used to continuously train the RL agent. In that way our RL designer may continually (online) learn through the behaviour of a human-like agent and adapt to the player's skill, preferences and even annotated experiences.

In this initial implementation of EDRL, we used a number of metrics to directly represent player experience in a theory-driven manner~\cite{yannakakis2018artificial}. In particular, we tested expressions of fun and historical deviation for generating segments of game levels. As future work, we intend to generate levels with adaptive levels of diversity, surprise~\cite{yannakakis2016searching,gravina2018quality}, and/or novelty and investigate additional proxies of player experience including difficulty-related functions stemming from the theory of flow~\cite{csikszentmihalyi2014flow}. Similarly to \cite{cideron2020qd,gravina2019procedural}, we are also interested in applying multi-objective optimisation procedures to consider simultaneously the quality and diversity of levels, instead of their linear aggregation with fixed weights. Viewing the EDRL framework from an intrinsic motivation \cite{barto2013intrinsic} or an artificial curiosity \cite{schmidhuber2006developmental} perspective is another research direction we consider. All above reward functions, current and future ones, will need to be cross-verified and tested against human players as in \cite{yannakakis2005ai}.

While EDRL builds on two general frameworks and it is expected to operate in games with dissimilar characteristics, our plan is to test the degrees to which EDRL is viable and scalable to more complex games that feature large game and action space representations. We argue that given a learned game content representation---such as latent vector of a GAN or an autoencoder---and a function that is able to segment the level, EDRL is directly applicable. Level design was our test-bed in this study; as both EDPCG and PCGRL (to a lesser degree) have explored their application to other forms of content beyond levels, EDRL also needs to be tested to other types of content independently or even in an orchestrated manner \cite{liapis2018orchestrating}. 

\section{Conclusion}\label{sec:conclusion}

In this paper, we introduced a novel framework that realises personalised online content generation by interweaving the EDPCG \cite{yannakakis2011experience} and the PCGRL \cite{khalifa2020pcgrl} frameworks. We test the ED(PRCG)RL framework, EDRL in short, in Super Mario Bros and train RL agents to design endless and playable levels that maximise notions of \emph{fun}~\cite{koster2013theory} and historical deviation. To realise endless generation in real-time, we employ a pre-trained GAN generator that designs level segments; the RL agent selects suitable segments (as represented by their latent vectors) to be concatenated to the existing level. The application of EDRL in SMB makes online level generation possible while ensuring certain degree of fun and deviation across level segments. The generated segments are automatically repaired by a  CNet-assisted Evolutionary Algorithm~\cite{shu2020cnet} and tested by an $A^*$ agent that guarantees playability. This initial study showcases the potential of EDRL in fast-paced games like SMB and opens new research horizons for realising experience-driven PCG though the RL paradigm. 

\section*{Acknowledgement}
The authors thank the reviewers for their careful reviews and insightful comments.

\balance
\bibliographystyle{IEEEtran}
\bibliography{main,dlpcg}
\end{document}